\documentclass[conference]{IEEEtran}
\IEEEoverridecommandlockouts
\usepackage[utf8]{inputenc}

\usepackage{cite}
\usepackage{amsmath,amssymb,amsfonts}
\usepackage{algorithmic}
\usepackage{textcomp}
\usepackage{graphicx}
\usepackage{xcolor}
\def\BibTeX{{\rm B\kern-.05em{\sc i\kern-.025em b}\kern-.08em
    T\kern-.1667em\lower.7ex\hbox{E}\kern-.125emX}}

\usepackage{hyperref}

\usepackage[acronyms,toc,nonumberlist,nogroupskip]{glossaries}
\glsdisablehyper
\glsnoexpandfields
\newacronym{ad}{AD}{automated driving}
\newacronym{bev}{BEV}{birds eye view}
\newacronym{ecdf}{ECDF}{empirical cumulative distribution function}
\newacronym{ooi}{OOI}{object of interest}
\newacronym{rss}{RSS}{responsibility sensitive safety}
\newacronym{ttc}{TTC}{time to collision}

\begin{document}

\title{Conservative Estimation of Perception Relevance of Dynamic Objects for Safe Trajectories in Automotive Scenarios\\

\thanks{The research leading to these results is funded by the German Federal Ministry for Economic Affairs and Climate Action within the project \emph{VVM - Verification Validation Methods} under grant number 19A19002S, as well as by the German Federal Ministry of Education and Research within the project \emph{VIVID} with the grant number 16ME0173 based on a decision of the Deutscher Bundestag.
The authors would like to thank the consortia for the successful cooperation.
}

}

\author{\IEEEauthorblockN{1\textsuperscript{st} Ken Mori}
\IEEEauthorblockA{\textit{Institute of Automotive Engineering} \\
\textit{Technical University Darmstadt}\\
Darmstadt, Germany \\
ken.mori@tu-darmstadt.de}
\and
\IEEEauthorblockN{2\textsuperscript{nd} Kai Storms}
\IEEEauthorblockA{\textit{Institute of Automotive Engineering} \\
\textit{Technical University Darmstadt}\\
Darmstadt, Germany \\
kai.storms@tu-darmstadt.de}
\and
\IEEEauthorblockN{3\textsuperscript{rd} Steven Peters}
\IEEEauthorblockA{\textit{Institute of Automotive Engineering} \\
\textit{Technical University Darmstadt}\\
Darmstadt, Germany \\
steven.peters@tu-darmstadt.de}
}


\IEEEoverridecommandlockouts
\IEEEpubid{\makebox[\columnwidth]{accepted for publication at IEEE Most'23~\copyright2023 IEEE \hfill} \hspace{\columnsep}\makebox[\columnwidth]{ }}

\maketitle

\IEEEpubidadjcol

\def\thefootnote{*}\footnotetext{The authors contributed equally to this work.}

\begin{abstract}
Having efficient testing strategies is a core challenge that needs to be overcome for the release of automated driving.
This necessitates clear requirements as well as suitable methods for testing. 
In this work, the requirements for perception modules are considered with respect to relevance.
The concept of relevance currently remains insufficiently defined and specified.

In this paper, we propose a novel methodology to overcome this challenge by exemplary application to collision safety in the highway domain.
Using this general system and use case specification, a corresponding concept for relevance is derived.
Irrelevant objects are thus defined as objects which do not limit the set of safe actions available to the ego vehicle under consideration of all uncertainties. 
As an initial step, the use case is decomposed into functional scenarios with respect to collision relevance.
For each functional scenario, possible actions of both the ego vehicle and any other dynamic object are formalized as equations.
This set of possible actions is constrained by traffic rules, yielding relevance criteria.

As a result, we present a conservative estimation which dynamic objects are relevant for perception and need to be considered for a complete evaluation.
The estimation provides requirements which are applicable for offline testing and validation of perception components. 
A visualization is presented for examples from the highD dataset, showing the plausibility of the results.
Finally, a possibility for a future validation of the presented relevance concept is outlined. 

\end{abstract}

\begin{IEEEkeywords}
Autonomous vehicles, Perception, Vision and scene understanding, Worst-case analysis
\end{IEEEkeywords}

\section{Introduction}
In recent years, \gls{ad} has been viewed as a key technology for improving societies' quality of life, including availability, efficiency as well as safety of travel~\cite{Canis.23.05.2017}. 
Due to the removal of human error from the traffic environment, a release of \gls{ad} has the potential to eliminate the cause of most traffic accidents~\cite{Canis.17.05.2018}.
However, the release of \gls{ad} also introduces the risk of new hazards stemming from errors in the automation itself~\cite{Wachenfeld.2016}.

For \gls{ad} to be released, it is necessary to prove that added benefits outweigh the new risks.
As such, it is critical to provide a safety argumentation, substantiated by evidence from testing, that a positive risk balance~\cite{UdoDiFabio.2017} is given.
Since current testing strategies are prohibitively expensive, new efficient approaches are required~\cite{Junietz.2018}.
One state of the art approach is to reduce effort of testing by modular decomposition~\cite{Amersbach.}.
A generally accepted first layer of modular decomposition of a driving function is dividing it into Sense, Plan and Act~\cite{philipp2020functional}.

The sense module includes sensing as well as perception~\cite{Amersbach.2020} and is tested detached from Plan and Act.
The requirements for safety can be defined by deriving them from known legal restrictions or normative target behavior~\cite{Salem.15.09.2022,Beck.20.07.2022}. 
However, such behavioral requirements are generally not applicable to the sense module. 
Therefore, perception components are generally underspecified~\cite{Spanfelner.2012, DAmour.2020}. 
Two aspects among the perception requirements are distinguished within this work: 
\begin{enumerate}
    \item What is relevant and needs to be perceived?
    \item How well does it need to be perceived?
\end{enumerate}

This work focuses on the first question and attempts to answer which objects are relevant for perception. 
Due to the complexity of specifying perception, only samples can be used as requirements~\cite{Spanfelner.2012}. 
In driving context, such samples are provided by perception datasets such as~\cite{Geiger.2012, Caesar.26.03.2019b}. 
While these datasets define relevance, this is only done implicitly by including or excluding objects from the ground truth.
Furthermore, the distinction is made based on arbitrary heuristics during labeling. 
However, this approach lacks consideration of safety~\cite{Willers.2020} or behavioral requirements. 

To address these shortcomings, this work presents a systematic method to define relevance for perception. 
The resulting requirements are intended for use in offline testing of perception. 

To maintain modularity, the concrete specification of the downstream Plan/Act module is considered unavailable. 
Therefore, the perception relevance is constructed to generalize across different planners and actors. 
To this end, the minimum behavioral requirements for the planner or the actor are considered. 
An object is considered relevant for the perception if the object necessitates an action according to the minimum behavioral requirements. 
We further explicitly consider uncertainties to obtain a conservative estimation of the perception relevance.

\section{Related Works}

Relevance is currently not sufficiently represented in existing perception metrics~\cite{Willers.2020}.
Therefore, this work will cover conceptualizations from different related domains.
While this work focuses on a nominal scale, relevance may be also be conceptualized on an ordinal scale~\cite{sperber_relevance_2001}.
Two key properties of relevance are for whom the information may be relevant~\cite{Schamber.1990} and the context in which the information is received~\cite{sperber_relevance_2001}.

\subsection{Relevance in Planning}

In previous work, deep learning with various different paradigms has been applied to the planning task~\cite{Huang.2020}.
Due to the black box nature of these systems, attention to the input of the networks is required. 
These inputs are typically filtered according to manually designed criteria.
Various works adopt a \gls{bev} grid with preset geometric boundaries of varying size as input to their planner~\cite{Bansal.07.12.2018, Sadat.2020, Philion.2020}. 
While not explicitly considered, the grid size defines a region of interest outside of which objects are implicitly declared irrelevant. 
Similar ideas also exist for the testing of \gls{ad} functions in simulation~\cite{Hallerbach.2018}.

A different approach is to directly consider the states of a fixed number of objects as planner input~\cite{Xu.07.12.2018, Cho.2019}.
Similar considerations are reflected in the Waymo Open Motion Dataset~\cite{Ettinger.2021} which limits validation and testing to a maximum of eight objects.
For evaluation, the Lyft Level 5 dataset~\cite{Houston.25.06.2020} focuses on single vehicles of interest and is not suited for multi-agent motion sequences~\cite{Vazquez.2022}.
In each case, objects beyond the preselected fixed number are declared irrelevant without explicit consideration.

A more explicit approach is given by formal models of planning, which attempt to assure specified safe planning behavior~\cite{Hoss.2022}.
An example is the \gls{rss} model, which formalizes the requirement of reasonable care to define proper responses to dangerous situations~\cite{ShalevShwartz.21.08.2017}.
Reachability analysis provides a different approach by defining sets of unsafe states with respect to an object, where probabilistic approaches may be used to avoid overly conservative estimations~\cite{Althoff.2010}.
Both approaches are formulated with respect to specific objects, which can be interpreted as relevant for the planning task.

\subsection{Standard Perception Evaluation}

One possible approach for defining relevance in perception is the application of heuristics~\cite{Hoss.2022, Berk.2020}.
A common example is the assumption that a lower distance to an object corresponds to a higher relevance~\cite{Lyssenko.2021}.

While relevance is rarely conceptualized explicitly on common dataset evaluation metrics, heuristic selection typically applies.
For example, KITTI only evaluates objects above a certain height when projected to the image plane~\cite{Geiger.2012} while nuScenes~\cite{Caesar.26.03.2019b} sets class-specific distance thresholds~\cite{nuScenes.2020}.
Other heuristics may be present from the annotation procedure.
For instance, nuScenes demands the presence of sensor detection points for object annotation~\cite{Caesar.26.03.2019b} while the Waymo Open Dataset limits the annotation range~\cite{Sun.10.12.2019}.

\subsection{Relevance for Perception Evaluation}

Considering object relevance for perception safety evaluation is necessary~\cite{Volk.2020} and has incorporated relevance according to the downstream task in two different ways. 
The first approach is to consider a concrete implementation of a planner to observe the effect of perturbations on the perception results~\cite{Philion.2020b, Philipp.2021, Henze.2021}.
In this case, the validity of the results obtained is restricted to the concrete planner~\cite{Philipp.2021}.

Other approaches consider more general specifications of the planner. 
One example is to incorporate modifications of \gls{rss} to distinguish relevant objects to define a safety metric~\cite{Volk.2020}.
Instead of a planner specification, reachability analysis relying on vehicle models have been applied to identify relevant vehicles~\cite{Topan.2022}.
Alternatively, behavioral requirements for different scenarios are considered for valet parking~\cite{Schonemann.2019} or an urban use case~\cite{Philipp.} 
In either case, either the behavioral requirements assume stopping to be valid behavior~\cite{Schonemann.2019, Topan.2022} or information on the ego intention and road environment are required~\cite{Philipp., Schonemann.2019}.

\subsection{Other Relevance Concepts}

This section discusses explicit considerations of relevance in visual scenes and for objects in other context than 3D perception.

In visual scenes, saliency conceptualizes the attempt to recognize important objects or regions in a scene~\cite{Ullah.2020}.
However, saliency suffers from subjectiveness and variance in annotation~\cite{Li.2015, Zhang.2020}.
A related approach is to consider eye fixation as proxy to identify informative regions~\cite{Chapman.1998, Ullah.2020}.
However, fixations depend on context and task~\cite{Chapman.1998, Makrigiorgos.2019} and neglect the ability of humans to interpret peripheral vision~\cite{Palazzi.10.05.2017}.
Additionally, it is unclear how to extend the concepts to 3D and objects which are occluded or not in the human field of view.

Closer relation to safety and 3D space is provided by the field of criticality metrics.
These metrics are typically used as a surrogate for safety in the process of safety validation~\cite{Junietz.2018b}.
A summary over many different criticality metrics is provided in~\cite{Mahmud.2017, Westhofen2021CriticalityMF}.
Many options beside spatial proximity exist and there is no clear consensus available on which metric and which threshold to apply.

\section{Abstract Method} \label{sec:AbstractMethod}

In this section, a novel abstract concept for conservative estimation of perception relevance based on collision safety is presented.

\begin{figure}
\includegraphics[width=\linewidth]{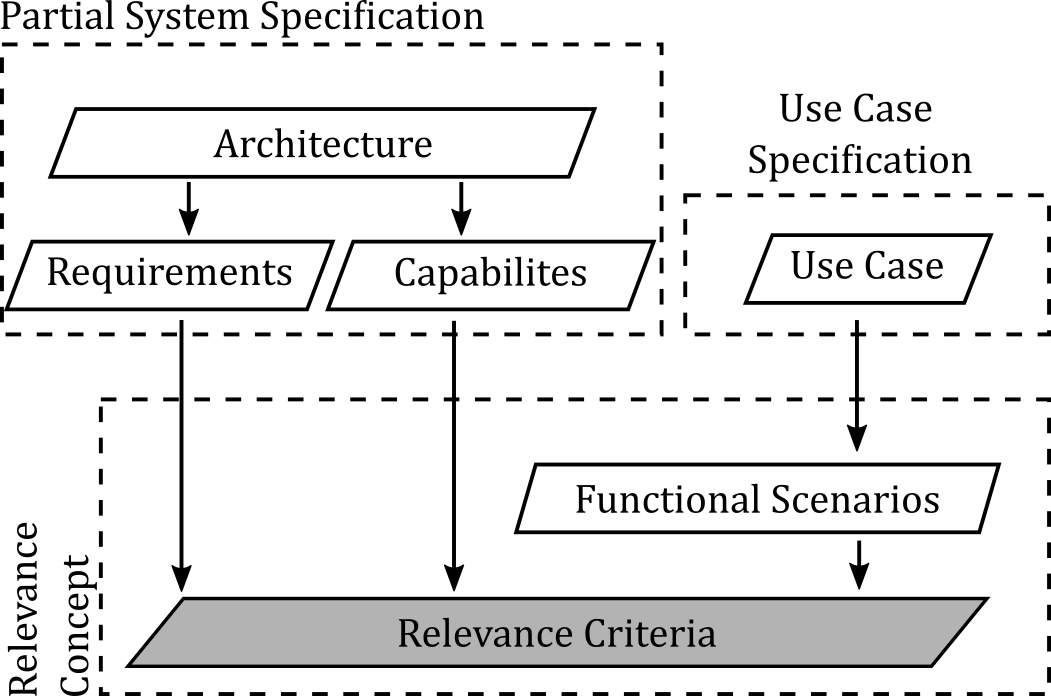} 
\caption{Overview of the proposed relevance method with its key outputs.}
\label{fig:process_relevance}
\end{figure}

In order to evaluate relevance, a definition is required first.
While different related terms are available in literature, this work considers situational relevance or utility, which is only well defined with respect to a task~\cite{Cosijn.2000}.
To simplify the terminology, this work continues to use the term relevance to refer to this concept. 
Thus, it is necessary to first define the task by specifying a system and its corresponding use case.
Based on this, the use case is decomposed into functional scenarios.
For each of the functional scenarios, the outer bounds of relevance are derived. 
The overall process is shown in Fig.\ref{fig:process_relevance}. 

\subsection{Partial System Specification} \label{sec:PartialSystemSpecification}

The partial system specification is required to later define and specify relevance. 
For this purpose, a generic architecture, a task including the high-level requirements as well as system capabilities are defined. 

\subsubsection{System Architecture} \label{sec:SystemArchitecture}
The partial specification is limited to the external interfaces between modules for the perception and downstream tasks.
The inner workings of each module remain a black box model.
For the system specification, a modular architecture composed of Sense, Plan and Act modules is assumed.
This simple architecture is commonly applied for \gls{ad} and serves as basis for other more complex architectures.
The sense module includes sensing as well as perception~\cite{Amersbach.2020}.
Firstly, a general object-centric output representation is selected.
An object list is assumed since it is the common representation between perception and planning~\cite{Hoss.2022}.
The downstream task is a plan module which outputs actions described by a trajectory.
These actions or trajectories are then executed by the act module.

\subsubsection{System Requirements} \label{sec:SystemRequirements}

To specify the task, a minimum set of system requirements for actions is defined.
Firstly, a valid planned trajectory must be executable by the downstream act module:
\begin{quote} \label{req:1}
REQ1: 
\emph{The actions must adhere to physical limitations.}
\end{quote}
Secondly, the actions must conform to traffic rules:
\begin{quote} \label{req:2}
REQ2: 
\emph{The actions must adhere to applicable legal restrictions.}
\end{quote}
These requirements are high-level and so far only applicable to the downstream planning task. 
For the further relevance concept, concrete requirements for the downstream task are required. 
To simplify the following considerations, the scope is limited to one concrete aspect. 
In this paper, we focus on the task of collision safety. 
These behavioral requirements are not directly applicable to the perception module.
How to link these requirements to perception for the definition of relevance is discussed in section~\ref{sec:RelevanceConcept}.

\subsubsection{System Capabilities}

Another aspect is the specification of the capabilities available to the system in order to fulfill these requirements. 
This work considers a system latency as well as available acceleration limits.

\begin{figure}
\includegraphics[width=\linewidth]{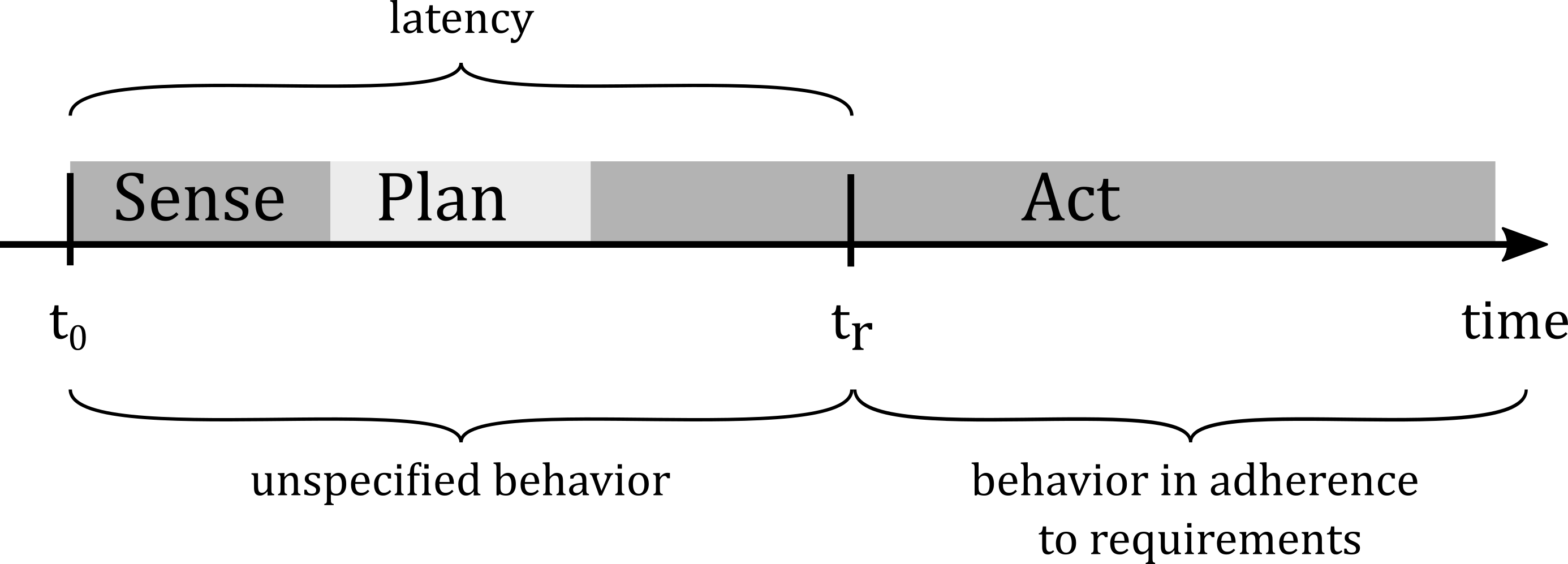} 
\caption{After the occurrence of an event, the ego behavior is unspecified with respect to this event. Only after a reaction time a behavior that incorporates the events behavioral requirements can be assumed.}
\label{fig:event_reaction}
\end{figure}

The latency in the system architecture is shown in Fig.~\ref{fig:event_reaction}.
Assume an event such as an action by another object occurs at the initial point in time \(t_\mathrm{0}\).
A certain time is required for the system to perceive the event, plan a trajectory and set its actuators to perform the planned trajectory.
The system latency or reaction time \(t_\mathrm{r}\) is the time from the event to the first actuator changing the state of the ego vehicle in response to the event. 
Therefore, the above requirements are not applied to the output of the plan module up to \(t_\mathrm{r}\).

Only after the reaction time, the system is assumed to act in accordance with the requirements. 
These capabilities may be limited due to outer circumstances or due to system specifications.
For simplicity, this work assumes that the system is capable of providing two different accelerations.

\begin{itemize}
    \item minimum guaranteed braking deceleration
    \item minimum guaranteed acceleration
\end{itemize}
The latter can refer either to accelerating in longitudinal direction or steering in lateral direction.
These general parameters describing the system capabilities are applied later in the relevance concept. 

\subsection{Use Case Specification} \label{sec:UseCaseSpecification} 

For the purpose of an initial presentation of the proposed methodology, a simple use case is considered in order to limit complexity.
Therefore, a highway environment is selected in this work.
Similar use cases are selected for example by \gls{rss}~\cite{ShalevShwartz.21.08.2017} for their initial development of ideas or in the PEGASUS project~\cite{PEGASUSProject.2019}.
Both a limited number of traffic participant types and a limited number of scenarios that may occur contribute to limiting the complexity.  

\subsection{Relevance Concept} \label{sec:RelevanceConcept}

Having defined both the system and the use case for the system, a corresponding relevance concept is defined.
The consideration of perception requirements as basis for relevance raises two core challenges.
First, these perception requirements are unknown or at least insufficiently specified.
Second, considerable uncertainties remain regarding aspects such as the future trajectories of other dynamic objects.
Additionally, reliable information for the road is unavailable and cannot be obtained without high effort.
This includes both high-level information such as the exact layout and low-level information like local friction coefficients. 
During the reaction time, the ego behavior with the resulting ego trajectory is not specified.

In order to overcome these challenges, we propose the following:
Instead of the the perception requirements, we first consider the acting requirements.
The requirements for acting are better specified, especially since they are subject to traffic rules.
The requirements for acting are each dependent on the context of the evaluated situation.
In order to consider this context dependence, the use case is decomposed into individual functional scenarios.
For each functional scenario, existing uncertainties are accounted for by assuming the worst case for each uncertainty in the given context. 

If the uncertainties are accounted for, this allows the transfer to perception requirements.
If a behavioral requirement with respect to an object exists, the object must be perceived. 
Therefore, the corresponding object is relevant for the perception task. 
Due to the uncertainties, the object is perception relevant if a potential behavioral requirement is possible within the uncertainties.
More specifically, if the worst case of all uncertainties results in a behavioral requirement, an object is relevant for perception.

\subsubsection{Use Case Decomposition}

Depending on different contexts within the use case, different behavioral requirements apply.
Decomposition of the use case into functional scenarios is used to specify the high-level requirements in section \ref{sec:SystemRequirements}. 
The decomposition results in identifiable functional scenarios with specific behavioral requirements.
This makes subsequent specification of relevance in the form of equations possible. 

The representation of the environment around the ego vehicle is assumed to be a generic object list as defined in the partial system specification.
Any object list limited to a given point in time contains many objects.
Similar to~\cite{Topan.2022}, only pairwise interactions between the ego and a single dynamic \gls{ooi} are considered.
This object's behavior may be additionally restricted by further objects.
Not considering further objects thus overestimates the possible behaviors, yielding conservative estimates while reducing complexity. 
Any object pair at a given point in time is defined by a parameter set including different values of parameters such as speed or distance.

For the use case, potential scenarios are identified. 
Similar approaches can be found in ~\cite{ShalevShwartz.21.08.2017, PEGASUSProject.2019, Philipp., Schonemann.2019}.
The conditions for which a functional scenario is applied are formalized in the form of equations.
These equations are wholly dependent on the given concrete parameter set of the object pair at the time of relevance evaluation.
Relying only on this parameter set, uncertainties exist regarding the road geometry and future behavior.
Depending on these factors, different scenarios such as a stopping or a merging procedure may unfold. 
Therefore, all applicable functional scenarios are considered as potential scenarios.
It needs to be noted that the scenarios are not mutually exclusive, meaning that multiple hypothetical scenarios may be considered.

For each hypothetical scenario, the relevance of the \gls{ooi} is evaluated as is described in the following section.
Similar to the superposition in~\cite{Schonemann.2019}, the object is considered relevant to the specified system if it exhibits relevance in any of the hypothetical scenarios.

\subsubsection{Relevance for Functional Scenarios}

This section outlines the suggested approach for defining relevance equations for a given functional scenario.
As first step, applicable behavioral requirements are extracted from legal requirements from German traffic law~\cite{BundesministeriumfurJustizundVerbraucherschutz.06.03.2013}.
The abstract requirements are then further specified and interpreted for the purpose of this work. 
The ego shows valid behavior after the specified system latency.
During the latency of the ego as well as throughout the scenario for the \gls{ooi}, worst case behavior is assumed. 

The respective behaviors are now specified as trajectories represented by parametrized equations. 
In order to develop these equations, simplified assumptions are used as in the worst time to collision metric~\cite{Wachenfeld.2016}. 
The objects are treated as point masses aside from the fact that they possess radii representing the object size. 
In addition, Kamm's circle is used as simple and comprehensive model to overapproximate the worst case action space of both vehicles. 
Kamm's circle assumes an isotropic maximum acceleration \(a_\mathrm{max}\) which is independent of the driving direction~\cite{Schmidt.2013}.
While kinematic constraints may further limit possible worst case actions, the assumptions represent conservative estimates as have been used in other work~\cite{Wachenfeld.2016, Althoff.2016}.

With these assumptions, it is possible to specify the respective possible behaviors of the dynamic objects as well as the solution space for the ego. 
Thus, the definition of relevance can be formulated as:
\begin{quote}
\emph{All objects that can change the set of viable trajectories are relevant for solving the combined planning/perception task.}
\end{quote}

\section{Method Application} \label{sec:MethodApplication}

After outlining the abstract approach, this section presents the application of said method.
This demonstrates the practicality of the method and allows comparison with heuristics from existing perception evaluation. 
The highway is chosen as the target domain due its presence in scientific literature.

\subsection{Use Case Decomposition} \label{sec:ScenarioDecompositionApplication}

This section describes how to decompose the previously defined use case into functional scenarios in order to obtain behavioral requirements. 
Common descriptions of scenarios such as by \gls{rss}~\cite{ShalevShwartz.21.08.2017} apply lane-based coordinates.
Contrary to \cite{ShalevShwartz.21.08.2017, Philipp., Schonemann.2019}, no reliable information on the lane geometry or the trajectories which may deviate from road lanes is assumed to be available. 
Therefore, different coordinates are required to represent the scenarios.

\begin{figure} 
\includegraphics[width=\linewidth]{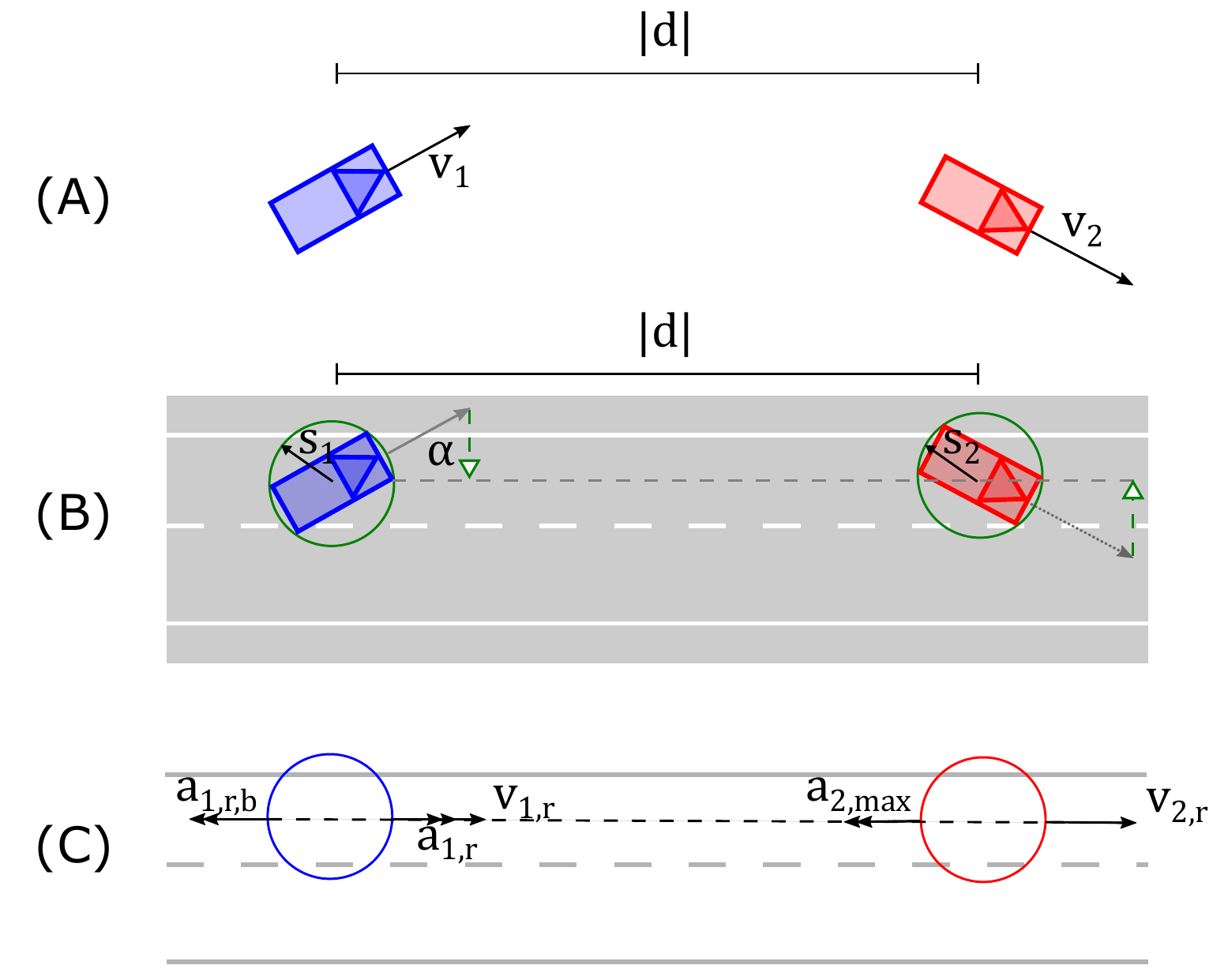} 
\caption{Development of a simplified environment model for a radial scenario. Shown for scenario R.TA. (A) Perceived real world (B) simplifications through hypothetical road model (C) simplified model )}
\label{fig:radial_relevance}
\end{figure}

The respective vehicle coordinates are unsuited, since implicitly a straight road in traveling direction is assumed. 
Polar coordinates are more reliable, since any collision requires the radial distance to approach zero. 
The situation is also visualized in Fig.~\ref{fig:radial_relevance} for the case where the road does not coincide with the direction in which the vehicles are travelling.

Based on the constellation of distance, radial and tangential velocity, different scenarios are distinguished.
If simplifications are required to derive equations, conservative estimates are used.
The following sections present the decomposition results first for radial and then for tangential scenarios.

\subsubsection{Radial Scenarios}

First, the different constellations which are possible with regards to the radial velocity are discussed. 
To formally describe the criteria to distinguish scenarios, the following notation is used.
Location vectors are denoted  by \(\vec{r}_\mathrm{i}\), where the index 1 refers to the ego vehicle while the index 2 refers to the \gls{ooi}.
This convention is maintained for all following equations with the first index referring to the vehicle if multiple indices are present.
Similarly, velocities are indicated by \(\vec{v}_\mathrm{i}\).
The vector connecting the vehicles \(\vec{d}\) is calculated as:
\begin{equation}
    \vec{d} = \vec{r}_\mathrm{2} - \vec{r}_\mathrm{1} 
\end{equation}

Generally, the following four pairings are possible.
The corresponding criteria for distinguishing the cases are also presented with the index 0 indicating the initial state.
\begin{itemize}
    \item ego moving towards \gls{ooi}, \gls{ooi} moving away from ego (R.TA): 
    \begin{equation}
        \vec{d}_\mathrm{0} \cdot \vec{v}_\mathrm{1,0} > 0 \ \cap \ 
        \vec{d}_\mathrm{0} \cdot \vec{v}_\mathrm{2,0} > 0
    \end{equation}
    \item ego moving away from \gls{ooi}, \gls{ooi} moving towards ego (R.AT): 
    \begin{equation}
        \vec{d}_\mathrm{0} \cdot \vec{v}_\mathrm{1,0} < 0 \ \cap \ 
        \vec{d}_\mathrm{0} \cdot \vec{v}_\mathrm{2,0} < 0
    \end{equation}
    \item both vehicles moving towards each other (R.TT): 
    \begin{equation}
        \vec{d}_\mathrm{0} \cdot \vec{v}_\mathrm{1,0} > 0 \ \cap \ 
        \vec{d}_\mathrm{0} \cdot \vec{v}_\mathrm{2,0} < 0
    \end{equation}
    \item both vehicles moving away from each other (R.AA): 
    \begin{equation}
        \vec{d}_\mathrm{0} \cdot \vec{v}_\mathrm{1,0} < 0 \ \cap \ 
        \vec{d}_\mathrm{0} \cdot \vec{v}_\mathrm{2,0} > 0
    \end{equation}
\end{itemize}

Each scenario is abbreviated with a three letter combination.
The first letter represents whether a radial or tangential scenario is considered.
The first letter after the period is whether the ego is moving towards or away from the object while the second letter conveys the same information for the \gls{ooi}.

Some simplifications which are applied to all of these scenarios are discussed in the following. 
The radial scenarios use a simple one-dimensional model.
All quantities are projected on the connecting line between the two vehicles in a conservative fashion. 
Hereby, the radial distance \(d\) between the objects is:  
\begin{equation}
    d = | \vec{d} |
\end{equation}
While the actual path may be longer, the direct distance is the shortest possible path and thus conservative. 
Tangential distance and velocity are neglected, resulting in:
\begin{equation}
    v_\mathrm{i,r} = | \vec{v}_\mathrm{i,r} | 
    = \left| \vec{v_\mathrm{i}} \cdot \frac{\vec{d}}{| \vec{d}|} \right|
\end{equation}
The second index r denotes the radial direction in this equation.
While this convention is maintained for all following equations, the index is optional and does not appear in all cases.
Again, this is conservative since tangential motion may prevent collisions by evasion.
Conservative estimates are used by assuming maximum acceleration to be available for the worst case behaviors of both vehicles.
The contractually guaranteed braking acceleration available to the ego in radial direction is reduced according to the direction of the ego velocity.
Part of the available acceleration due to friction may be required for tangential motion, i.e.\ to follow a curved road.
This limits the acceleration available for braking in radial direction \(a_\mathrm{i,r,b}\) to:
\begin{equation}
    a_\mathrm{i,r,b} = cos(\alpha) \cdot a_\mathrm{i,b} 
    = \frac{v_\mathrm{i,r}}{v_\mathrm{i}} \cdot a_\mathrm{i,b}
\end{equation}
In this case, the last index b refers to the braking procedure.
The optional last index refers to states or events in all following equations.

Acceleration and steering are limited by the vehicle~\cite{Bokare.2017} and human preference~\cite{Bertolazzi.2010} rather than the available friction on the road.
Therefore, the available guaranteed acceleration is not reduced so that \(a_\mathrm{i,r,g} = a_\mathrm{i,g}\). 

\subsubsection{Tangential Scenarios}

For the tangential direction, only two cases require distinction.
\begin{itemize}
    \item \gls{ooi} moving away from ego (T.XA):
    \begin{equation}
        \vec{d}_\mathrm{0} \cdot \vec{v}_\mathrm{2,0} \ge 0
    \end{equation}
    \item \gls{ooi} moving towards ego (T.XT):
    \begin{equation}
        \vec{d}_\mathrm{0} \cdot \vec{v}_\mathrm{2,0} < 0
    \end{equation}
\end{itemize}

\begin{figure}
\includegraphics[width=\linewidth]{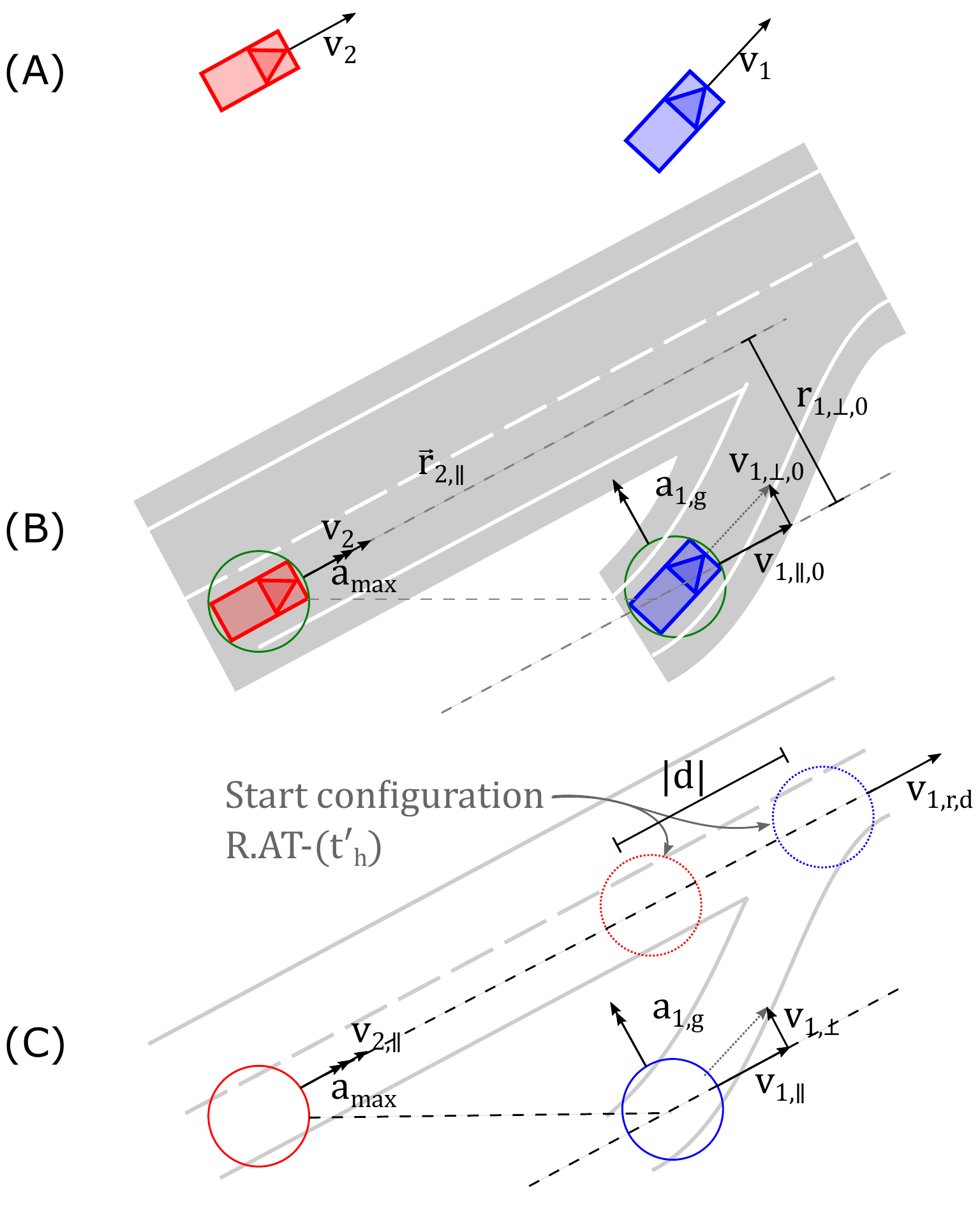} 
\caption{Development of a simplified environment model for a tangential scenario. Scenario T.XT is shown. (A) Perceived real world (B) simplifications through hypothetical road model (C) simplified model}
\label{fig:tangential_relevance}
\end{figure}

The abbreviations are used as previously with the distinction that the X signifies that the ego motion direction is not relevant. 

If the \gls{ooi} is moving away, tangential dynamics are irrelevant.
It should be noted that neglecting tangential dynamics does not mean the absence of relevance since the radial scenarios are considered separately.
If the \gls{ooi} is moving towards the ego, there exists a potential merging scenario where the ego merges in front of the \gls{ooi} as depicted in Fig.~\ref{fig:tangential_relevance}.

\subsection{Relevance for Functional Scenarios}

This section elaborates the application of the previously explained method to derive equations for each functional scenario defined in the previous section. 

\subsubsection{R.TA: Ego moving towards \gls{ooi}, \gls{ooi} moving away from ego}

This scenario corresponds to the case that the ego is following the \gls{ooi}.
For this case, German traffic regulations (StVO) state the following requirement~\cite[p.3]{BundesministeriumfurJustizundVerbraucherschutz.06.03.2013}:
\begin{quote}
REQ2.1: 
    \emph{The ego vehicle shall be able to brake to halt behind a vehicle in front to avoid a collision in the event that the front car suddenly brakes.}
\end{quote}

The same notion has been demanded as the rule to not hit others from behind as part of the \gls{rss} model~\cite[p.6-7]{ShalevShwartz.21.08.2017}.
This requirement of collision avoidance applies even for worst case assumptions and can be formalized for the minimum distance \(d_\mathrm{min}\) as:
\begin{equation} \label{eq:NonCollision}
    d_\mathrm{min} > 0
\end{equation}

For the one-dimensional model, the behavior of the vehicles is given as acceleration in radial direction.
The worst case behavior during the latency is the ego vehicle accelerating towards the \gls{ooi}.
Throughout the scenario, the \gls{ooi} has a braking acceleration directed in the direction of the ego vehicle. 

The worst case accelerations are \(a_\mathrm{max}\), while the valid reaction of the ego is undertaken with the contractually specified \(a_\mathrm{1,b}\).
Note that for the case of braking, negative acceleration values apply.
The position \(r_\mathrm{i}\) and velocity \(v_\mathrm{i}\) for a constant acceleration are given by:

\begin{equation} \label{eq:Position}
    r_\mathrm{i,r} =  r_\mathrm{i,r,0} + v_\mathrm{i,r,0} \ t + \frac{1}{2} a_\mathrm{i,r,0}  t^2 
\end{equation} 
\begin{equation} \label{eq:Velocity}
    v_\mathrm{i,r} =  v_\mathrm{i,r,0} + a_\mathrm{i,r,0} \ t
\end{equation} 
The first index again denotes the vehicle while the second index 0 refers to the initial state.
For both vehicles, the corresponding radial braking distance \(r_\mathrm{i,r,b}\) is:

\begin{equation} \label{eq:BrakingDistance}
    r_\mathrm{i,r,b} = \frac{v_\mathrm{i,r}^2}{2 a_\mathrm{i,r,b}}
\end{equation}

The ego vehicle accelerates during the reaction time and then brakes until coming to a full stop. 
Considering its initial acceleration in \eqref{eq:Position} and for the braking \eqref{eq:BrakingDistance} by applying \eqref{eq:Velocity}, its position after coming to a full stop is:

\begin{equation} \label{eq:EgoReactBrake}
\begin{aligned}
    r_\mathrm{1,r,s} =  & \ r_\mathrm{1,r,0} + v_\mathrm{1,r,0} \ t_\mathrm{1,r} + \frac{1}{2} a_\mathrm{1,r,0}  t_\mathrm{1,r}^2 \\
    &+ \frac{(v_\mathrm{1,r,0} + t_\mathrm{1,r} a_\mathrm{1,r,0})^2}{2 a_\mathrm{1,r,b}} 
\end{aligned}
\end{equation} 

The variable \(t_\mathrm{1,r}\) denotes the reaction time of the ego vehicle. 
The minimal distance is achieved when both vehicles have come to a full stop.
This corresponds to subtracting \eqref{eq:BrakingDistance} from the braking distance of the \gls{ooi} specified according to \eqref{eq:Position}.
Additionally, the sizes of both vehicles \(s_\mathrm{i}\) are subtracted similar to \cite{Wachenfeld.2016}.
Including the worst case assumptions, radial components and \eqref{eq:NonCollision} yields:
\begin{equation} \label{eq:FollowingDistance}
\begin{aligned}
    0 < d_\mathrm{min} 
    = & \  d_\mathrm{0} - s_\mathrm{1} - s_\mathrm{2} + \frac{v_\mathrm{2,r,0}^2}{2a_\mathrm{max}}  - v_\mathrm{1,r,0} \ t_\mathrm{1,r} \\
    &- \frac{1}{2} a_\mathrm{max}  t_\mathrm{1,r}^2 
    - \frac{(v_\mathrm{1,r,0} + t_\mathrm{1,r} a_\mathrm{max})^2}{2 a_\mathrm{1,r,b}}
\end{aligned}
\end{equation} 
If this condition is fulfilled, no collision is available for any worst case behavior. 
If this condition is violated, collision avoidance may potentially restrict the available ego actions, which means the other object is relevant.

\subsubsection{R.AT: ego moving away from \gls{ooi}, \gls{ooi} moving towards ego} \label{R.AT}

This case corresponds to the situation where the ego vehicle is being followed by another vehicle. 
The behavioral requirement from German law is that~\cite{BundesministeriumfurJustizundVerbraucherschutz.06.03.2013}: 
\begin{quote}
REQ2.2: \label{req:2.2}
    \emph{other vehicles should not be unnecessarily impeded.}
\end{quote}

Within this work, impeding is interpreted to mean that the ego vehicle exerts additional action requirements regarding collision avoidance on the \gls{ooi}.
This case is conceptually the same as exchanging the roles in the R.TA scenario.
The \gls{ooi} therefore has the same requirements of not causing collisions with the ego vehicle. 
It is possible to distinguish the case where the ego vehicle speed is adequate (R.AT+) and the case where the ego speed is below the desired traveling speed (R.AT-).

\subsubsection{R.AT+: ego with desired speed moving away from \gls{ooi}, \gls{ooi} moving towards ego}

For this scenario it is assumed that the speed chosen by the \gls{ooi} is adequate. 
In this case, \hyperref[req:2.2]{REQ2.2} can be interpreted as the following subrequirement: 
\begin{quote}
REQ2.3: 
    \emph{The ego vehicle may not restrict the actions of the following vehicle by unnecessarily braking.}
\end{quote}
Exchanging the roles in the R.TA scenario allows reusing the results of the previous section:

\begin{equation} \label{eq:DistanceMinRAT+}
\begin{aligned}
    0 < d_\mathrm{min} 
    = & \  d_\mathrm{0} - s_\mathrm{1} - s_\mathrm{2} + \frac{v_\mathrm{1,r,0}^2}{2a_\mathrm{max}} - v_\mathrm{2,r,0} \ t_\mathrm{2,r} \\
    & - \frac{1}{2} a_\mathrm{max}  t_\mathrm{2,r}^2 
    - \frac{(v_\mathrm{2,r,0} + t_\mathrm{2,r} a_\mathrm{max})^2}{2 a_\mathrm{2,r,b}}
\end{aligned}
\end{equation} 
Note that the reaction time \(t_\mathrm{2,r}\) and the guaranteed braking acceleration \(a_\mathrm{2,r,b}\) refer to the \gls{ooi} in this case. 
The equation then signifies the all cases where a full braking of the ego vehicle does not yet restrict the action space of the following \gls{ooi}.
Violating the equation means that the ego restricts the action space of the \gls{ooi} and thus hinders it which is only permitted if a valid reason is present.
Therefore, the other object is relevant.

\subsubsection{R.AT-: ego with less than desired speed moving away from \gls{ooi}, \gls{ooi} moving towards ego}

For this case, it is assumed that the ego speed is initially lower than the desired ego speed.
This situation may occur for example after a lane change onto a faster lane. 
In this case, the ego vehicle first attempts to accelerate to the desired speed \(v_\mathrm{1,r,d}\).
After that, the scenario is identical to the R.AT+ scenario.
Until the requirements of R.AT+ apply, \hyperref[req:2.2]{REQ2.2} can be interpreted as:
\begin{quote}
REQ2.4: 
    \emph{The ego vehicle may not restrict the actions of the following vehicle by having insufficient speed.}
\end{quote}

The worst case requirements in this case do not consider the ego latency, since this is later taken into account when the lane change is initiated. 
Therefore, the ego vehicle simply accelerates towards its desired speed.
The \gls{ooi} accelerates towards the ego vehicle throughout the scenario.

Using \eqref{eq:Position} and the worst case assumptions, the distance between the two vehicles can be computed as:

\begin{equation} \label{eq:DistanceRAT-}
    d = d_\mathrm{0} - s_\mathrm{1} - s_\mathrm{2} + (v_\mathrm{1,r,0}-v_\mathrm{2,r,0}) t + \frac{1}{2} (a_\mathrm{1,r,g}-a_\mathrm{max}) t^2
\end{equation}
The critical situation is the case where the minimal distance is reached. 
Since the available guaranteed ego acceleration \(a_\mathrm{1,g}\) is generally smaller than the worst case acceleration \(a_\mathrm{max}\) assumed to be available for the \gls{ooi}, the minimum distance occurs when the ego vehicle reaches the desired speed indicated by the index d.
The desired velocity is unknown in the absence of knowledge about the speed limit, but can be conservatively estimated to be the velocity of the \gls{ooi}. 

\begin{equation}  \label{eq:TimeDesired}
    t_\mathrm{d} = \frac{v_\mathrm{1,r,d} - v_\mathrm{1,r,0}}{a_\mathrm{1,r,g}}
\end{equation}
Inserting \eqref{eq:TimeDesired} into \eqref{eq:DistanceRAT-} and \eqref{eq:Velocity} yields the distance between the vehicles and the velocity of the \gls{ooi} at the time \(t_\mathrm{d}\), respectively.
With \eqref{eq:Velocity}, the corresponding velocity of the \gls{ooi} is:

\begin{equation} \label{eq:TimeDesiredVelocity}
    v_\mathrm{2,r,d} = v_\mathrm{2,r,0} + a_\mathrm{max} t_\mathrm{d}
\end{equation}
Equation \eqref{eq:DistanceMinRAT+} from the R.AT+ scenario is adapted with the following modifications.
The initial distance considering object sizes \((d_\mathrm{0}-s_\mathrm{1} - s_\mathrm{2})\) is substituted with the result obtained from inserting \eqref{eq:TimeDesired} into \eqref{eq:DistanceRAT-}.
The initial ego velocity \(v_\mathrm{1,r,0}\) is substituted by the desired velocity \(v_\mathrm{1,r,d}\) and \eqref{eq:TimeDesiredVelocity} is used instead of \(v_\mathrm{2,r,0}\), resulting in:

\begin{equation} \label{eq:DistanceMinRAT-}
\begin{aligned}
    0 < d_\mathrm{min} 
    = & \  d(t = t_\mathrm{d}) + \frac{v_\mathrm{1,r,d}^2}{2a_\mathrm{max}} 
    - v_\mathrm{2,r,d} \ t_\mathrm{2,r} \\
    & - \frac{1}{2} a_\mathrm{max}  t_\mathrm{2,r}^2 
    - \frac{(v_\mathrm{2,r,d} + t_\mathrm{2,r} a_\mathrm{max})^2}{2 a_\mathrm{2,r,b}}
\end{aligned}
\end{equation} 
Note that the reaction time \(t_\mathrm{2,r}\) and guaranteed braking acceleration \(a_\mathrm{2,r,b}\) refer to the \gls{ooi} in this case. 
As with the other scenarios, objects violating the requirement are considered relevant.

\subsubsection{R.TT: both vehicles moving towards each other}

This case may occur if vehicles are travelling in lanes on opposite sides of the road or if they are moving laterally towards each other. 
For two vehicles moving towards each other, \gls{rss} assumes a correct reaction from both vehicles~\cite{ShalevShwartz.21.08.2017}.
A similar understanding is reflected in the StVO which demands that stopping should be possible within half of the visible distance~\cite{BundesministeriumfurJustizundVerbraucherschutz.06.03.2013}.
However, this does not align with worst case assumptions which are only given by the physically possible dynamic limits of the dynamic objects~\cite{Wachenfeld.2016}.
In this scenario, the worst case for the ego vehicle during the reaction time is accelerating towards the \gls{ooi}.
After the reaction time, it begins braking while the \gls{ooi} accelerates towards the ego vehicle throughout the scenario.

For this case, the legal requirement of avoiding to harm or endanger others~\cite{BundesministeriumfurJustizundVerbraucherschutz.06.03.2013} requires further interpretation to extract behavioral requirements.

If accidents are unavoidable, minimizing harm and damage is reasonable.
While the ego vehicle cannot influence the other dynamic objects, it can minimize its own contribution to the damage by minimizing its speed.
Therefore, the requirement is:
\begin{quote} \label{req:2.5}
REQ2.5: 
    \emph{the ego vehicle shall brake to a standstill before the other vehicle collides with it.}
\end{quote}
This means that the requirement for the minimum distance \(d_\mathrm{min}\) is: 
\begin{equation} \label{eq:NonCollisionRTT}
    d_\mathrm{min} > 0
\end{equation}

This requirement should be fulfilled for the worst case braking maneuver of the ego vehicle as described by \eqref{eq:BrakingDistance}.
The velocity before and after the reaction time is described as follows:
\begin{equation} \label{eq:VeloBeforeReact}
    v_\mathrm{1,r} = v_\mathrm{1,0} + a_\mathrm{max} t \mathrm{\quad for \quad} t \le t_\mathrm{r}
\end{equation}
\begin{equation} \label{eq:VeloAfterReact}
    v_\mathrm{1,r} = v_\mathrm{1,r,b} - a_\mathrm{1,r,b} (t - t_\mathrm{r})  \mathrm{\quad for \quad} t \ge t_\mathrm{r}
\end{equation}
In this case, \(v_\mathrm{1,r,b}\) is the velocity of the ego vehicle after the reaction time when it begins to brake. 
The ego braking time \(t_\mathrm{1,b}\) is obtained by inserting \eqref{eq:VeloBeforeReact} into \eqref{eq:VeloAfterReact} and demanding that \(v_\mathrm{1,r} = 0\):

\begin{equation}
    t_\mathrm{1,b} = t_\mathrm{1,r} + \frac{(v_\mathrm{1,r,0} + t_\mathrm{1,r} \ a_\mathrm{max})}{a_\mathrm{1,r,b}}
\end{equation}
The position of the \gls{ooi} is described by \eqref{eq:Position}, but the velocity and the acceleration are negative in this scenario.
From this, the ego position after coming to a full stop in \eqref{eq:EgoReactBrake} and the sizes of the vehicles are subtracted.
Inserting the worst case assumptions as well as \eqref{eq:NonCollisionRTT} yields:

\begin{equation} \label{eq:DistanceRTT}
\begin{aligned}
    0 < d_\mathrm{min} = 
    & \ d_\mathrm{0} - s_\mathrm{1} - s_\mathrm{2} - v_\mathrm{1,r,0} \ t_\mathrm{1,r} \\
    & - \frac{1}{2} a_\mathrm{max}  t_\mathrm{1,r}^2 - \frac{(v_\mathrm{1,r,0} + t_\mathrm{1,r} a_\mathrm{max})^2}{2 a_\mathrm{1,r,b}} \\
    & - v_\mathrm{2,r,0} t_\mathrm{1,b} - \frac{1}{2} a_\mathrm{max} t_\mathrm{1,b}
\end{aligned}
\end{equation}
The velocities of the vehicles are considered to be positive if they are aimed towards each other.
Objects violating this equation are objects for which a collision is possible and which are thus relevant.

\subsubsection{R.AA: both vehicles moving away from each other}

This scenario may again correspond to vehicles travelling on opposite lanes or to vehicles laterally moving towards each others.
Especially for the latter case, it becomes clear that an ego braking may be required if the \gls{ooi} reverses its moving direction.
This case is fundamentally the same requirement as \hyperref[req:2.5]{req:2.5} as described in the previous section.
Since the velocities of the vehicles are pointed away from each other, negative velocities are inserted in \eqref{eq:DistanceRTT}.
If the requirement is violated, a collision is possible and the object is therefore considered relevant.

\subsubsection{T.XT: \gls{ooi} moving towards ego} \label{seq:T.XT}

This scenario corresponds to a potential merging scenario where the ego vehicle attempts to merge in front of the \gls{ooi}. 
Similarly to previous sections, the behavioral requirement \hyperref[req:2.2]{REQ2.2} that other vehicles should not be unnecessarily impeded applies~\cite{BundesministeriumfurJustizundVerbraucherschutz.06.03.2013}.
The ego vehicle is assumed to move onto the other lane and then accelerate.
For this scenario, it is assumed that there is a suitable target speed for the lane onto which the \gls{ooi} is moving.
The ego vehicle must not impede the \gls{ooi} before reaching the target speed as specified in the R.AT- scenario. 
Once the target speed has been reached the scenario becomes the previously treated R.AT+ scenario.

For this scenario, lane based coordinates with lateral and longitudinal direction are more suitable than the previously applied polar coordinates. 
Since no lane information is assumed available, a hypothetical worst case lane is constructed. 
This worst case lane is given by the current direction of movement of the \gls{ooi} since this allows the \gls{ooi} to accelerate towards the ego without requiring acceleration for change of direction.
The longitudinal direction \(r_\mathrm{i,\parallel}\) is given by: 

\begin{equation}\vec{r}_\mathrm{i,\parallel} = \begin{bmatrix} x_\mathrm{i,\parallel} \\ y_\mathrm{i,\parallel} \end{bmatrix} =
    \frac{\vec{v}_\mathrm{i}}{|\vec{v}_\mathrm{i}|} 
\end{equation}
The lateral direction \(\vec{r}_\mathrm{i,\bot}\) can be defined by switching the vector components of \(\vec{r}_\mathrm{i,\parallel}\) with each other:

\begin{equation}
    \vec{r}_\mathrm{i,\bot} = \begin{bmatrix} x_\mathrm{i,\bot} \\ y_\mathrm{i,\bot} \end{bmatrix} 
    = \begin{bmatrix} - y_\mathrm{i,\parallel} \\ x_\mathrm{i,\parallel} \end{bmatrix} 
\end{equation}
With this, the two velocity components of the ego vehicle perpendicular and parallel to the direction of the movement of the \gls{ooi} can be calculated according to: 

\begin{equation}
    v_\mathrm{1,\bot} = \vec{v}_\mathrm{1} \cdot \vec{r}_\mathrm{2,\bot}
    \mathrm{\quad and \quad}
    v_\mathrm{1,\parallel} = \vec{v}_\mathrm{1} \cdot \vec{r}_\mathrm{2,\parallel} 
\end{equation}

The movement of the ego vehicle is modeled as a process consisting of two parts.
First, the lateral movement corresponding to the lane change is performed without longitudinal acceleration. 
Afterwards, the ego vehicle accelerates exclusively in longitudinal direction.
This assumption is conservative since a combined steering and accelerating may potentially shorten the merging procedure.

During the lane change, worst case assumptions for the ego latency are required. 
However, initially steering in the wrong direction prior to a lane change in the opposite direction is unrealistic.
Moreover, the lane change itself is generally the worst case scenario since traveling or braking on the current lane must be possible to ensure safety. 
This means that in case of initial movement in the wrong direction, the lane change can be aborted.
Therefore, it is instead assumed that the lateral velocity \(v_\mathrm{1,\bot}\) is directed towards the \gls{ooi}:

\begin{equation}
    (\vec{d}_\mathrm{0} \cdot \vec{r}_\mathrm{2,\bot}) \cdot \vec{v}_\mathrm{1,0} \ge 0
\end{equation}
If this condition is violated, the scenario is not considered to be a potential merging scenario. 
Afterwards, the ego vehicle begins accelerating in lateral direction towards the hypothetical lane.
It then decelerates so that it reaches a lateral velocity of \(v_\mathrm{1,\bot} = 0\) at the lateral position \(r_\mathrm{1,\bot}=0\).
Throughout the scenario, the \gls{ooi} simply accelerates along its direction of movement. 
After that the scenario can be treated as the R.AT- scenario where the ego is required to accelerate until reaching the desired target speed. 

The lateral velocity and position during the reaction time \(t_\mathrm{r}\) are:
\begin{equation} \label{eq:VLatReact}
    v_\mathrm{1,\bot} = v_\mathrm{1,\bot,0} + a_\mathrm{max} \cdot t 
\end{equation}
\begin{equation} \label{eq:SLatReact}
    r_\mathrm{1,\bot} = r_\mathrm{1,\bot,0} + v_\mathrm{1,\bot,0} \cdot t + \frac{1}{2} a_\mathrm{max} t^2 
\end{equation}
To obtain the values at the end of the reaction time, it is necessary to consider the fact that the lateral velocity is required to be negative. 
Therefore, it is necessary to consider the time where a lateral of velocity of \(v_\mathrm{1,\bot,s} = 0\) is reached, if this case occurs, by inserting the following in the equations above:  
\begin{equation}
    t_\mathrm{1,r}' = min \biggl\{ t_\mathrm{1,r} \ , \ - \frac{v_\mathrm{1,\bot,0 }}{a_\mathrm{max}} \biggr \}
\end{equation}

After the reaction time, the ego vehicle performs a deliberate lateral movement to change lanes. 
For brevity, the substitution \(t' = t - t_\mathrm{r} \) is introduced.
The index s indicates the point in time after the reaction time of the ego where it begins switching lanes.
The index c is used for the point in time where the ego vehicle changes its behavior from laterally accelerating to laterally decelerating with the same acceleration of \(a_\mathrm{1,g}\) since both are steering maneuvers.
The movement is described by:

\begin{equation} \label{eq:VLatAccel}
    v_\mathrm{1,\bot} = v_\mathrm{1,\bot,s} - a_\mathrm{1,g} \cdot t' \mathrm{\quad for \quad} t' \le t_\mathrm{c}'
\end{equation}
\begin{equation}\label{eq:VLatBrake}
    v_\mathrm{1,\bot} = v_\mathrm{1,\bot,c} + a_\mathrm{1,g} \cdot (t' - t_\mathrm{c}') \mathrm{\quad for \quad} t' > t_\mathrm{c}'
\end{equation}
The corresponding location in lateral direction considering the available acceleration of \(a_\mathrm{g}\) for \(t' \le t_\mathrm{c}'\) is described by:

\begin{equation} \label{eq:SLatAccel}
    r_\mathrm{1,\bot} = r_\mathrm{1,\bot,s} + v_\mathrm{1,\bot,s} \cdot t' - \frac{1}{2} a_\mathrm{1,g} t'^2 
\end{equation}
After the reaction time, the location for \(t' > t_\mathrm{c}'\) is:

\begin{equation} \label{eq:SLatBrake}
    r_\mathrm{1,\bot} = r_\mathrm{1,\bot,c} + v_\mathrm{1,\bot,c} \cdot (t' - t_\mathrm{c}') 
    + \frac{1}{2} a_\mathrm{1,g} (t' - t_\mathrm{c}')^2 
\end{equation}
The ego vehicle comes to a stop in lateral direction at the desired position \(r_\mathrm{1,\bot} = 0\) after a halting time \(t_\mathrm{h}'\).
These conditions are given by:

\begin{equation} \label{eq:VLatCondition}
    v_\mathrm{1,\bot,h} = v_\mathrm{1,\bot}(t=t'_\mathrm{h}) = 0
\end{equation}
\begin{equation} \label{eq:SLatCondition}
    r_\mathrm{1,\bot,h} = r_\mathrm{1,\bot}(t=t'_\mathrm{h}) = 0
\end{equation}
By solving this equation system, the time required for the lateral movement is calculated.
First, \eqref{eq:VLatAccel} and \eqref{eq:VLatBrake} are inserted into \eqref{eq:VLatCondition}: 

\begin{equation}
    0 = v_\mathrm{1,\bot,h}
    = v_\mathrm{1,\bot,s} + a_\mathrm{1,g} \cdot (t'_\mathrm{h} - 2 t'_\mathrm{c} )
\end{equation}
Next, \eqref{eq:VLatReact}, \eqref{eq:SLatReact} and \eqref{eq:SLatBrake} are inserted into \eqref{eq:SLatCondition}.

\begin{equation}
\begin{aligned}
    0 = & \ r_\mathrm{1,\bot,s} + v_\mathrm{1,\bot,s} \cdot t'_\mathrm{c} - \frac{1}{2} a_\mathrm{1,g} {t'_\mathrm{c}}^2 \\
    & + (v_\mathrm{1,\bot,s} - a_\mathrm{1,g} \cdot t_\mathrm{c}) \cdot (t'_\mathrm{h} - t'_\mathrm{c}) \\
    & + \frac{1}{2} a_\mathrm{1,g} (t'_\mathrm{h}-t'_\mathrm{c})^2 
\end{aligned}
\end{equation}

\noindent 
These equations can be solved for \(t_\mathrm{h}\), thus yielding the time required for the lateral movement.
\begin{equation}
    t'_\mathrm{c} = \frac{v_\mathrm{1,\bot,s}}{a_\mathrm{max}} + \sqrt{ \frac{r_\mathrm{1,\bot,s}}{a_\mathrm{1,\bot,g}} }
\end{equation}
\begin{equation}
    t'_\mathrm{h} = t'_\mathrm{c} - \frac{v_\mathrm{1,\bot,s}}{a_\mathrm{1,\bot,g}}
\end{equation}

During the lateral movement, a constant longitudinal velocity is assumed.
After completing the lateral movement, the ego vehicle accelerates as in scenario R.AT-.
However, instead of the radial direction everything is projected to the longitudinal direction.
In addition, the time to reach the desired velocity from \eqref{eq:TimeDesired} must be modified to consider the reaction time as well as the time required for the lateral movement:

\begin{equation} \label{eq:TimeTXT}
    t'_\mathrm{d} = t_\mathrm{r} + t'_\mathrm{h} + \frac{v_\mathrm{1,\parallel,d} - v_\mathrm{1,\parallel,0}}{a_\mathrm{1,\parallel,g}}
\end{equation}

\noindent
As previously, the desired velocity is unknown in the absence of knowledge about speed restrictions.
However, it can be conservatively estimated by using the speed of the \gls{ooi} so that \(v_\mathrm{1,\parallel,d} = v_\mathrm{2,0}\).
Modifying \eqref{eq:TimeDesiredVelocity} and \eqref{eq:DistanceRAT-} accordingly yields:

\begin{equation}
    v'_\mathrm{2,d} = v_\mathrm{2,0} + a_\mathrm{max} t'_\mathrm{d}
\end{equation}
\begin{equation} \label{eq:DistanceDesiredTXT}
\begin{aligned}
    d(t = t'_\mathrm{d}) = & \ r_\mathrm{1,\parallel,0} - s_\mathrm{1} - s_\mathrm{2} + (v_\mathrm{1,\parallel,0}-v_\mathrm{2,0}) t'_\mathrm{d} \\ & + \frac{1}{2} (a_\mathrm{1,\parallel,g}-a_\mathrm{max}) {t'_\mathrm{d}}^2
\end{aligned} 
\end{equation}
Substituting for \(t_\mathrm{d}\) and \(v_\mathrm{2,r,d}\) in \eqref{eq:DistanceMinRAT-} delivers:

\begin{equation} \label{eq:DistanceTXT}
\begin{aligned}
    0 < d_\mathrm{min} 
    = & \  d(t = t'_\mathrm{d}) + \frac{v_\mathrm{1,\parallel,d}^2}{2a_\mathrm{max}} 
    - v'_\mathrm{2,d} \ t_\mathrm{r} \\
    & - \frac{1}{2} a_\mathrm{max}  t_\mathrm{r}^2 
    - \frac{(v'_\mathrm{2,d} + t_\mathrm{r} a_\mathrm{max})^2}{2 a_\mathrm{1,\parallel,g}}
\end{aligned}
\end{equation} 
Any objects violating this requirement are considered relevant for the ego vehicle.

\section{Results}

\begin{figure*}
  \centering
  
\includegraphics[width=\textwidth]{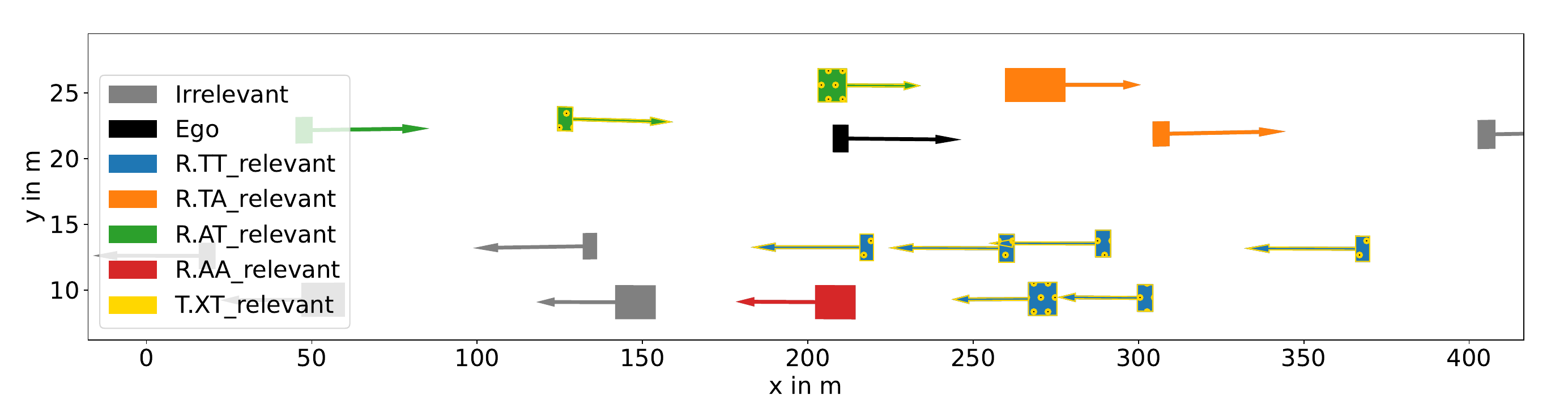}

\caption{Visualizations of object relevance resulting from different functional scenarios on a highway segment from the highD dataset. Since tangential and radial relevance criteria may overlap, tangential relevance is indicated by the yellow edge and dotted hatch.}
\label{fig:visualization_highD}
\end{figure*}

This section presents results from applying the presented method to publicly available datasets.
Popular perception datasets typically focus on urban scenarios and only cover a limited sensor range which may limit their suitability for highway applications~\cite{Wang.2021}.
Instead, this work presents results on the highD dataset~\cite{Krajewski.2018}. 
The highD dataset includes vehicle trajectories extracted from video recordings from a drone.
Vehicle trajectories are available for German highways and encompass road segments with a length of approximately 420~m. 

In order to apply the equations developed in previous sections, concrete parameter values are required. 
The values are chosen to be realistic under typical conditions encountered on a highway.
However, they may not be applicable to any and all situations and further substantiation of the argument is required. 
The maximum acceleration limited by available friction is chosen as \(a_\mathrm{max} = 10 \ \frac{m}{s^2}\) in accordance with~\cite{Althoff.2016, Wachenfeld.2016}. 
Braking deceleration during emergency stops reach at minimum \(a_\mathrm{b} = 7 \ \frac{m}{s^2}\) even on wet road surfaces~\cite{PoulGreibe.2007}.
The available longitudinal acceleration is speed dependant and differs for different types of vehicles.
An acceleration of \(a_\mathrm{g} = 0.5 \ \frac{m}{s^2}\) is chosen based on results from~\cite{Bokare.2017}.
For lateral maneuvers, drivers generally display accelerations which are smaller than the physically feasible accelerations.
At highway speeds, the accelerations are between \(1-2 \ \frac{m}{s^2}\), indicating that the guaranteed acceleration is limited by the longitudinal acceleration.
For the reaction times, a common value for humans in the case of surprise intrusions of 1.5~s is found~\cite{Green.2000}.
These values are used for both of the vehicles where applicable.

The method of this work is only applied to a subset of the highD data to limit computation and memory consumption.
Only the first of multiple track files from the dataset is used, since these 348 000 bounding boxes are considered sufficient for this work. 
Results are visualized in two different ways.
A exemplary overview for a single frame is presented in Fig.~\ref{fig:visualization_highD} to provide intuitive results.

\begin{figure}
\includegraphics[width=\linewidth]{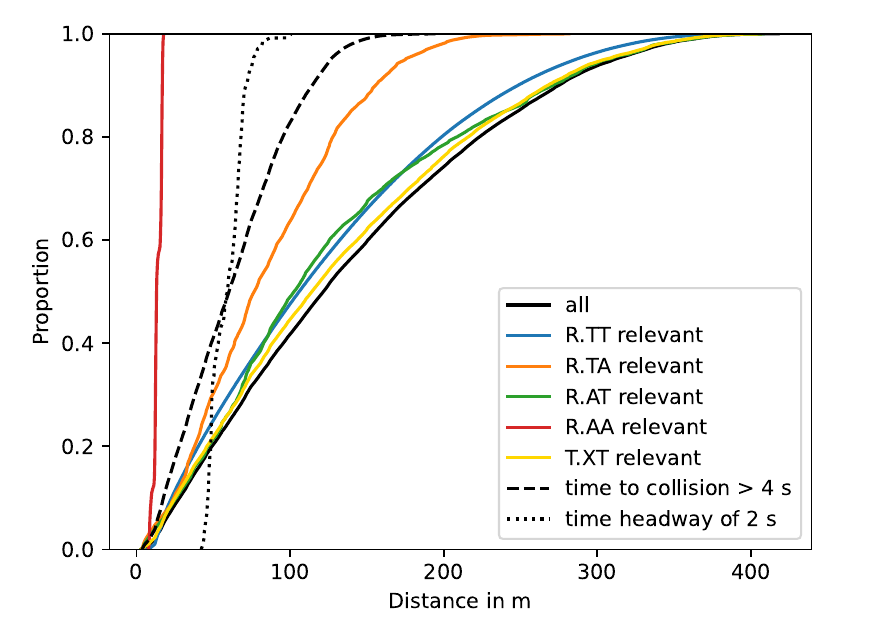} 
\caption{\gls{ecdf} of distances for all objects and objects which are considered relevant according to different criteria. The distribution for a typical suggested time gap for driving on a highway is provided as reference.}
\label{fig:distribution_highD}
\end{figure}

Note that the aspect ratio is not equal in order to improve visibility especially for the lateral velocity components. 
More detailed information regarding the distribution of relevant objects according to different criteria is provided in Fig.~\ref{fig:distribution_highD}.
Each line represents an \gls{ecdf} over the distance for a given category of relevance or objects.
Additional reference is provided by visualizing the two common criticality metrics time headway and \gls{ttc}.
The common recommendation of road administrations is a time headway of 2~s~\cite{Mahmud.2017}.
By multiplying the time headway requirement with the ego velocity, the results can be directly displayed as distance distribution.
\gls{ttc} requirements in literature range from 1.5-4 s depending on source and situation~\cite{Mahmud.2017}.
Since this work only uses the values to provide an intuitive reference, the threshold is arbitrarily selected to be a upper value of 4~s. 
By only considering objects below the \gls{ttc} threshold, another distance distribution is obtained.

\section{Discussion}

This work first presents a general method for conceptualizing and deriving relevance. 
Its practicality was shown by applying it to the use case of object detection on highways. 
Nevertheless, we consider the method sufficiently general to be applicable to other systems and use cases. 
Examples are different systems such as a traffic sign detector or a different use case such as urban environments. 
However, the procedure must be modified to consider different requirements and scenarios. 

The application of the method to the use case yields the relevance of objects. 
Overall, the exemplary results appear plausible. 
In addition, they are supported by an argumentation which considers safety. 
Since all results are based on conservative estimates, the process is designed to avoid false negatives regarding relevance. 
The distance distributions in Fig.~\ref{fig:distribution_highD} are mostly above the distances indicated by time headway and time to collision criteria. 
Since the other criteria are not designed for the R.AA scenario, this scenario is not considered here. 
While the time headway intersects the relevance criteria of this work, the \gls{ttc} indicates that this is due to its lack of consideration of relative velocity. 
Therefore, the conservative design yields the expected results. 
On the other hand, this increases the likelihood of false positives regarding relevance. 
However, in the context of testing perception this is the intended behavior. 
Safety is improved since stricter performance requirements are imposed on the perception systems.

One further question is whether the method is applicable to other road regulations than the German StVO~\cite{BundesministeriumfurJustizundVerbraucherschutz.06.03.2013} used as basis in this work.
While details differ, examples of other traffic rules such as the California Driver's Handbook~\cite{StateofCaliforniaDepartmentofMotorVehicles.2022} and the Japanese Road Traffic Act~\cite{MinistryofJustice.1960} share similar general objectives of avoiding collisions and ensuring safety. 
Since collision safety is generally only implicitly specified, this work proposed to further specify safety while considering high-level objectives.
Thus, specifics of the German StVO are only weakly expressed in this work.
While further explicit validation is required, a transfer to other traffic rules seems plausible. 

The parametrization of the equations is only required for the final application to the real world dataset. 
Generally, the parameter values depend on the use case as well as the system. 
Therefore, the assumed reaction times and braking accelerations for the \gls{ooi} may not apply to all traffic participants and situations.
Conditions such an icy or oily road surface do not occur in the considered dataset.
Such circumstances would further reduce the available deceleration and acceleration.
Thus, in other applications such conditions need to be considered in choosing the parameter values or an argumentation for their omission is required. 

Considering these requirements developed in this work has implications for perception testing. 
One notable fact is that the distance of the objects at which they become relevant depends on the scenario.
This indicates that heuristics of circular regions of relevance as in~\cite{Geiger.2012, Caesar.26.03.2019b} are unsuited to the highway domain.
Overall, the distances obtained in this work are large, reaching approximately 250 m for the R.TA scenario and up to 400 m in the other scenarios. 
Notably, this value is substantially higher than sensor range of 250 m by the Cirrus dataset~\cite{Wang.2021} already explicitly targeting highway applications.
Furthermore, the values approach the boundaries imposed by the highD dataset used for visualization.
This may indicate that more explicit consideration of relevance for perception and a corresponding argumentation are required. 

The results of this work allow to identify objects which are relevant to the perception task. 
However, an overall validation of the results is difficult since to the best of our knowledge, no validation methods are available in literature. 
Nevertheless, it should be noted that current perception evaluation implicitly defines relevance. 
The predominant approach is using arbitrary heuristics using distance or human visibility.
Therefore, the principled argumentation presented in this work is considered a substantial improvement.

\section{Conclusion \& Outlook}

A concept for perception specific relevance and a novel approach for determining a conservative estimation of perception relevance was presented.
The method was applied to the use case of highway driving. 
It was subsequently evaluated and discussed based on existing data of German highways.
The results indicate that the heuristics applied in current datasets may be insufficient for perception testing.
Future work may attempt to integrate the results of this work into datasets for perception testing.
An additional potential use case is the design of sensor setups. 
Leveraging the information on relevance may aid the decision what sensors and what detection ranges are required. 

Within the scope of this paper, the use case was limited to highways as proof of concept.
The authors consider the application to the systems and use cases as a possible next step in research. 
Examples include urban environments or other systems such as a traffic sign detection. 
This may yield insights regarding the general applicability of the presented method. 
However, the requirements, scenarios and equations may differ substantially.
Extending the presented method accordingly and studying the differences is left for future work.

Another aspect for future research is the validation of the methods presented.
The authors propose an approach relying on evaluating an prediction component.
We consider this an adaptation of the fundamental approach presented by \cite{Philion.2020b}.
As a modification, it is possible to substitute the planner with a prediction component trained on human driving behavior.
One advantage of prediction over planning is open loop training.
In addition, the objective for training and evaluation is better specified.
The overall performance of the prediction component is expected to deliver the reliability of the prediction for validation in the given scenario. 
By contrasting different inputs for the prediction, it is possible to obtain an estimation if the relevance definition is adequate. 
The prediction is first applied with all objects including both relevant and irrelevant objects as input. 
Here, the prediction quality can directly be assessed if offline data is available. 
Next, the prediction is applied with only the relevant objects as input.
If the relevant objects are correctly determined, the prediction performances in both cases are expected to be identical.
A drop in prediction performance indicates an object falsely identified as irrelevant.
The specificity of the relevance model may be assessed by comparing the prediction performance with all relevant objects against a subset thereof.
If the prediction performance remains unchanged, it may be indicative that an irrelevant object was falsely declared relevant.

We hope that this work encourages a more explicit consideration of relevance for perception evaluation and can serve as a future baseline.


\bibliographystyle{IEEEtran}
\bibliography{Mo, St}

\end{document}